\documentclass[runningheads]{llncs}
\usepackage[T1]{fontenc}
\usepackage{graphicx,verbatim}
\usepackage{amssymb, amsmath}
\usepackage{marvosym}
\begin{document}
\title{CTSL: Codebook-based Temporal-Spatial Learning for Accurate Non-Contrast Cardiac Risk Prediction Using Cine MRIs}
%

\titlerunning{CTSL: Codebook-based Temporal-Spatial Learning}
%
\author{Haoyang Su\inst{1,2,4} \and
Shaohao Rui\inst{2,3,4} \and
Jinyi Xiang\inst{3} \and
Lianming Wu\inst{3} \and
Xiaosong Wang\inst{4}\textsuperscript{(\Letter)}}

\authorrunning{H. Su et al.}
%
\institute{Fudan University, Shanghai, China \and Shanghai Innovation Institute, Shanghai, China  \and
Shanghai Jiao Tong University, Shanghai, China \and Shanghai Artificial Intelligence Laboratory, Shanghai, China\\
\email{wangxiaosong@pjlab.org.cn}}
\maketitle              
\begin{abstract}
Accurate and contrast-free Major Adverse Cardiac Events (MACE) prediction from Cine MRI sequences remains a critical challenge. Existing methods typically necessitate supervised learning based on human-refined masks in the ventricular myocardium, which become impractical without contrast agents. We introduce a self-supervised framework, namely Codebook-based Temporal-Spatial Learning (CTSL),  that learns dynamic, spatiotemporal representations from raw Cine data without requiring segmentation masks. CTSL decouples temporal and spatial features through a multi-view distillation strategy, where the teacher model processes multiple Cine views, and the student model learns from reduced-dimensional Cine-SA sequences. By leveraging codebook-based feature representations and dynamic lesion self-detection through motion cues, CTSL captures intricate temporal dependencies and motion patterns. High-confidence MACE risk predictions are achieved through our model, providing a rapid, non-invasive solution for cardiac risk assessment that outperforms traditional contrast-dependent methods, thereby enabling timely and accessible heart disease diagnosis in clinical settings.


\end{abstract}

\section{Introduction}
The application of MACE in survival analysis within cardiology is of paramount importance, serving as a critical indicator of long-term cardiac health and treatment outcomes~\cite{Bosco2021MajorAC,Revista2022Survival,RAZIPOUR2025AI}. In this context, Cine cardiac MRI imaging is widely accessible, while its prognostic efficacy is significantly hindered by the inherent intricacy of myocardial tissue and the entanglement of its temporal and spatial dynamics~\cite{Rajiah2023Cardiac}. Classical methods~\cite{Cox2018CoxPH,Katzman2016DeepSurvPT,Nagpal2021DSM}, modeled through electronic health records (EHR) or radiomics, purely rely on manual interpretations of structural and functional abnormalities~\cite{BANIECKI2025Interpretable,Bello2018DeepLC}, which are subject to inter-observer variability and often fail to capture subtle, yet crucial, prognostic features. 
Though the landscape of state-of-the-art survival models for 3D medical imaging is vast, limitations still persist. XSurv~\cite{Meng2023MergingDivergingHT}, which utilizes multi-modal data such as PET and CT scans, struggles with the scarcity of paired samples and the challenges of data co-registration. AdaMSS~\cite{Meng2024AdaMSS}, which requires physician-driven lesion refinement, is both time-consuming and labor-intensive. Furthermore,  models specialized in pathology~\cite{Zhao2023SelfSupervised,Ramanathan2024Ensemble,Jaume2024Modeling,Hou2023Multi} are limited by their reliance on 2D imaging, resulting in poor generalization to high-dimensional images. As a result, while Cine imaging is a commonly available modality, its integration of multi-dimensional data, including multi-chamber dynamics from short-axis and longitudinal views of cardiac morphology over time, still remains a challenge in survival analysis.

In this work, we first present a self-supervised pre-training scheme, denoted as CTSL, which operates independently of heart masks or contrast imaging data.  The framework mainly comprises two stages: motion-aware multi-view model distillation and spatiotemporal disentangling. Initially, we extend the classical distillation learning paradigm, DINOv2~\cite{oquab2024dinov2learningrobustvisual}, from a patient-level perspective, innovatively incorporating multi-view Cine sequences as input for the distillation, i.e., injecting the information from other views than short-axis (SA) images into the pre-trained model. In this stage, motion queries extracted through SA Cine sequences are treated as myocardium-oriented key tokens by the student network, which aligns with long-axis Cine tokens from the teacher network via  Kullback-Leibler (KL) divergence~\cite{Kullback1951OnIA}. Subsequently, drawing upon the latent space discretization techniques of VQVAE~\cite{oord2018neuraldiscreterepresentationlearning}, we extract query tokens from the preceding KL-aligned student model and design trainable temporal and spatial codebook embeddings, disentangling the spatiotemporal representations from the compressed 4D Cine data. 
Finally, a survival prediction framework is presented using the learned image tokens from CTSL and EHR features to perform MACE-based survival analysis.

Our contributions in the proposed framework are threefold: 1) We demonstrate the feasibility of adopting contrast-free imaging techniques together with EHR for the MACE survival analysis.  2) We introduce a self-supervised framework, CTSL, that learns codebook-based spatiotemporal representations from raw Cine data via a motion-aware multi-view model distillation module and a spatiotemporal feature disentanglement module. 3) We evaluate the proposed survival analysis framework on three private datasets and demonstrate its superior performance compared to prior arts.

\section{Method}
\begin{figure}[t]
    \centering
    \includegraphics[width=1\linewidth]{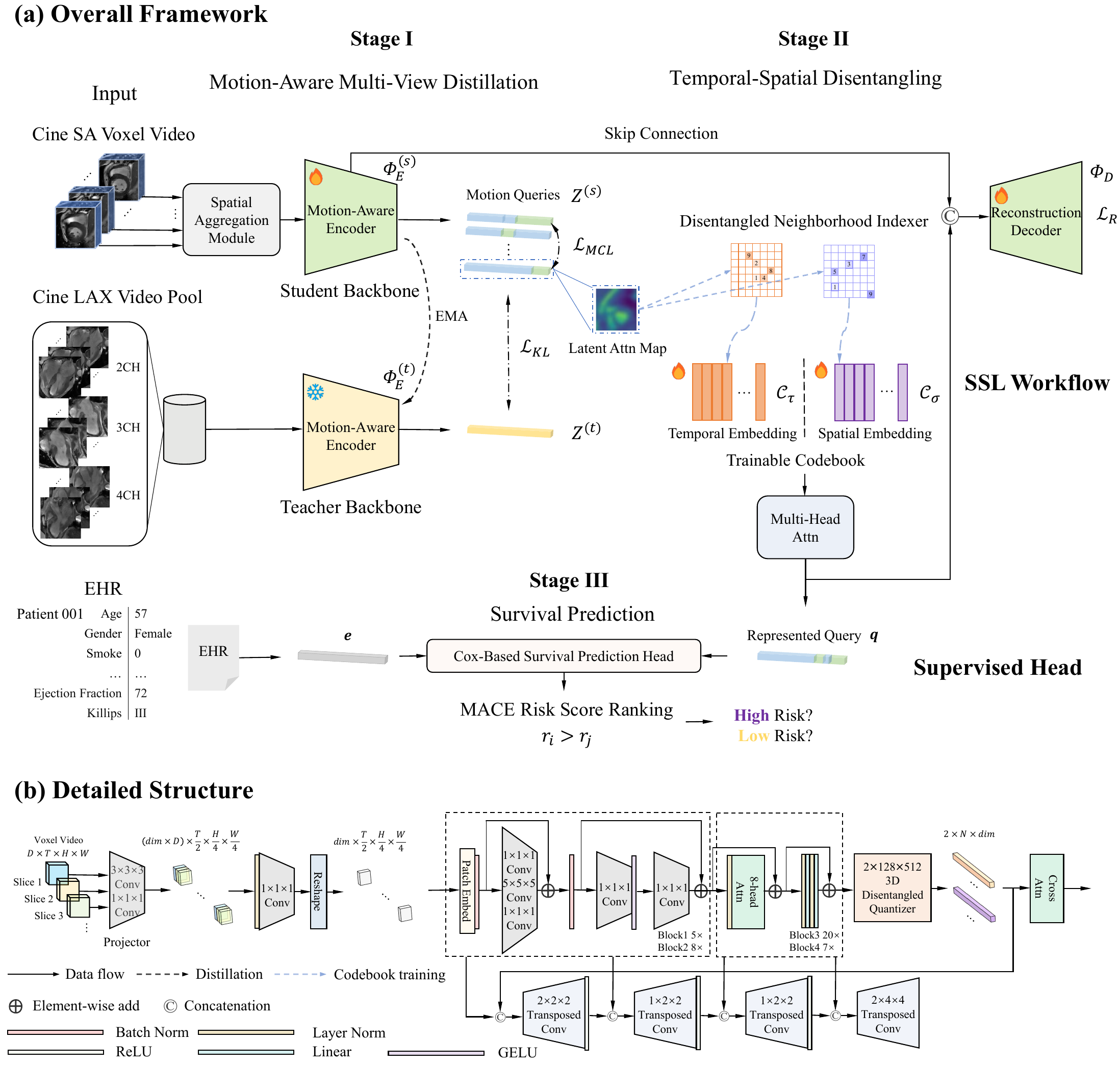}
    \caption{Overall framework and detailed structure of CTSL. (a) Self-supervised cardiac risk prediction framework. (b) Model architecture with  Uniformer~\cite{li2022uniformer} backbone.}
    \label{fig:framework_CTSL}
\end{figure}

We propose the CTSL framework shown in  Fig.~\ref{fig:framework_CTSL}, which operates through a two-stage self-supervised learning paradigm, followed by a final survival prediction stage. In stage I, multi-view cine sequences are processed independently by teacher and student networks, where KL loss $\mathcal{L}_{KL}$ aligns SA dynamics with long-axis anatomical patterns, while motion-aware distillation is enforced via motion contrastive loss $\mathcal{L}_{MCL}$. Spatiotemporal codebooks are composed based on student-derived features using nearest-neighbor indexing, generating compact representations for survival prediction. At last, final represented queries, fused with EHR data, drive Cox-based risk stratification through parameterized temporal and spatial embeddings.

\subsection{Preprocessing: Adaptive Myocardial Motion Localization}
We apply a mask-free Region-of-Interest (ROI) preprocessing strategy,
where optical flow is employed to extract the myocardial motion-focused region $\mathcal{V}$ from the full heart Voxel Video $\mathcal{V}^{(total)}\in \mathbb{R}^{H^{(total)}\times W^{(total)}\times T\times D}$.
The Farneback dense optical flow algorithm~\cite{Farnebck2003TwoFrameME} is applied to estimate the motion field $\mathbf{F}^{(t)}$ between adjacent Cine frame slices.
\begin{equation}
    \mathbf{F}^{(t)}=\Psi_{\mathrm{FB}}\Big(\mathcal{I}_t, \mathcal{I}_{t+1}\Big) \in \mathbb{R}^{H \times W \times 2}, ~\forall t \in \{0, \dots, T-1\}.
\end{equation}

The global-level ROI center $\bar{\mathbf{c}}$ is determined by aggregating the centroid trajectories across time windows. A window width of $s=96$ is utilized as the resolution of the ROI, obtaining the resulting Cine myocardial voxel video to be fed into the SSL framework as $\mathcal{V} = \mathcal{V}^{(total)}[\bar{c_y}-s/2:\bar{c_y}+s/2, \bar{c_x}-s/2:\bar{c_x}+s/2, :, :] \in \mathbb{R}^{H \times W \times T \times D}$.

\subsection{Stage I: Motion-aware Multi-view Model Distillation}
Given the preprocessed 4D Cine ROI sequence $\mathcal{V}\in \mathbb{R}^{H\times W\times T\times D}$, we designed paired motion-aware encoders $\Phi_E^{(s)}$ and $\Phi_E^{(t)}$ through a teacher-student distillation paradigm. A spatial aggregation module $\Gamma_p$ first processes the input through depth-wise feature extraction and obtain
\begin{equation}
    \Gamma_p (\mathcal{V})=\mathrm{concat}\Big[\Phi_p^{(d)}(\mathcal{V}_{:,:,:,d})\Big],
\end{equation}
where each depth-specific operator $\Phi_p^{(d)}:\mathbb{R}^{1\times T\times H\times W}\rightarrow \mathbb{R}^{64\times \frac{T}{2}\times \frac{H}{4}\times \frac{W}{4}}$ implements temporal-dominant 3D convolutions. The concatenation operator concat[$\cdot$]
preserves motion patterns across depth dimensions, which further yields $Z_{0}$.

The classical video architecture Uniformer~\cite{li2022uniformer} is employed as the backbone, where we extract the pre-logits motion queries $Z^{(s)}=\Phi_E^{(s)}(Z_0^{SA})$ from the student network's Cine SA input, while aggregating the teacher network's long-axis features as  $Z^{(t)}=\Phi_E^{(t)}([Z_0^{CH_2},Z_0^{CH_3},Z_0^{CH_4}])$.

To reconcile motion disparity and enforce patient-level alignment, we formulate a hybrid loss to enable the teacher network to be updated through Exponential Moving Average (EMA),
\begin{equation}
\mathcal{L}_{\mathrm{StageI}} =\tau^2 D_{\mathrm{KL}}\left(p^{(s)}| \bar{p}^{(t)}\right) + \lambda\mathbb{E}\left[\log\frac{\exp(\langle \mathbf{z}_i^{(s)},\mathbf{z}_i^{(s,+)}\rangle/\tau_c)}{\sum\limits_{j}\exp(\langle \mathbf{z}_i^{(s)},\mathbf{z}_j^{(s,-)}\rangle/\tau_c)}\right],
\end{equation}
where $p^{(s)}=\mathrm{Softmax}(Z^{(s)}/\tau), \bar{p}^{(t)}=\mathrm{Softmax}(Z^{(t)}/\tau)$, and $\mathbf{z}_i^{(s)}=\frac{Z_i^{(s)}}{||Z_i^{(s)}||_2}$.
The distillation term minimizes KL divergence between student predictions and teacher ensembles, while the contrastive term aligns SA features $\mathbf{z}_i^{(s)}$ with temporally synchronized positives $\mathbf{z}_i^{(s, +)}$ and repulsing negatives $\mathbf{z}_j^{(s, -)}$ from other patients. This dual mechanism leverages anatomical consistency, strictly aligning motion trajectory, especially at the End-Diastole and End-Systole phases.

\subsection{Stage II: Spatiotemporal Codebook Learning with Disentangled Representation}

The temporal and spatial motion queries $Z^{(s)}_\tau$ and $Z_\sigma^{(s)}$ are obtained through the trained encoder $\Phi_E^{(s)}$ in Stage I. Disentangled spatiotemporal representations through vector quantization are constructed. Let $\mathcal{C}_\tau, \mathcal{C}_\sigma\in\mathbb{R}^{n_e\times d_c}$
denote trainable temporal and spatial with $n_e=128$ and $d_c=\mathrm{dim}=512$, respectively.

For each codebook entry $\mathrm{e}_k\in\mathcal{C} = \{ \mathcal{C}_\tau, \mathcal{C}_\sigma\}$, quantized embeddings $Q=\{Q_\tau, Q_\sigma\}$ are obtained through
\begin{equation}
    Q = \mathrm{VecQuant}(Z^{(s)}, \mathcal{C})=\sum\limits_{k=1}^{n_e}\mathbb{I}\Big(k=\mathrm{argmin_j}||Z^{(s)}-\mathbf{e}_j||_2^2\Big)\mathbf{e}_k.
\end{equation}
Represented query $Q_{img}$ is obtained by cross attention through spatiotemporal quantization interaction,
\begin{equation}
    Q_{img} = \mathrm{Softmax}\Big(\dfrac{Q_\tau Q_\sigma}{\sqrt{d_c}}\Big)Q_\sigma\in \mathbb{R}^{N_\tau\times d_c}.
\end{equation}
The joint optimization objective integrates codebook learning with spatiotemporal reconstruction
\begin{equation}
\mathcal{L}_{\text{StageII}} = ||\Phi_D(Q_{\mathrm{img}}, {Z_l^{(s)}) - \mathcal{V}||_2^2} + \alpha\Big[||Z_\tau^{(s)} - \text{sg}(Q_\tau)||_2^2 + ||Z_\sigma^{(s)} - \text{sg}(Q_\sigma)||_2^2\Big],
\end{equation}
where sg($\cdot$) denotes stop-gradient, $\Phi_D$ denotes the reconstruction decoder, and $\alpha$ balance loss components. The dual codebook design $(\mathcal{C}_\tau, \mathcal{C}_\sigma)$ encourages resolving ambiguities where temporal blurring obscures spatial boundaries, such as the confusion between trabeculations and papillary muscles~\cite{CHUANG2012Correlation,INAGE2015IMPACTS}, which often occurs in entangled encoding paradigms.

\subsection{Survival Prediction Head}
Clinical biomarkers $\mathbf{e}\in\mathbb{R}^{d_m}$ contained in tabular EHR data are exploited to formulate the multimodal fusion head with refined image features $\mathbf{q}=\mathbb{E}(Q_{img})\in\mathbb{R}^{d_c}$. The fusion features $x_{\mathrm{fused}}=\mathrm{concat}[\mathbf{e}, \mathbf{q}]\in \mathbb{R}^{d_m+d_c}$ are then obtained.

A classical Cox head is defined, whose coefficients $\beta_k$ automatically weight cross-modal interactions. The hazard function is subsequently defined as
\begin{equation}
    h(t|\mathbf{x}_{\mathrm{fused}})=h_0(t)\mathrm{exp}\Big(\sum\limits_k\beta_kx_{\mathrm{fused}_k}\Big).
\end{equation}
The loss function minimizes negative log partial likelihood:
\begin{equation}
    \mathcal{L}_{\mathrm{Cox}}=-\sum_{i:\delta_i=1}\Big[\mathbf{\theta^{\mathrm{T}}x^{(i)}_{\mathrm{fused}}}-\mathrm{log}\sum_{j\in R(t_i)}\mathrm{exp}(\mathbf{\theta^{\mathrm{T}}}\mathbf{x}^{(j)}_{\mathrm{fused}})\Big]+\lambda||\mathbf{\theta}||_2^2,
\end{equation}
where $\mathbf{\theta}=[\beta_1,...,\beta_m]^{\mathrm{T}}$, $\delta_i\in\{0,1\}$ indicates event occurrence, and $R(t_i)$ is the risk set at time $t_i$.

\section{Experiments}
\noindent\textbf{Datasets.}
Three in-house cardiac CINE MRI datasets, i.e., RJCCM, AZCCM, and TJCCM, were utilized in the experiments. Each set includes four standardized views: short-axis, 2-chamber, 3-chamber, and 4-chamber orientations, comprising 407, 673, and 313 studies from patients, along with matched EHR data containing 135, 173, and 74 cardiovascular risk factors, respectively.

All sequences from the three datasets apply a magnetic field strength of 3.0 T and a 16-bit allocated intensity resolution for each image. Protocols across acquisition sites are variable. RJCCM
employed a system with repetition time (TR) = 2.95-3.02\;ms and echo time (TE) = 1.45-1.5\;ms, capturing 30-phase cardiac cycles; AZCCM utilized a system with TR = 12.4-13.5\;ms and TE = 1.55-1.61\;ms with 25-phase cardiac cycles; TJCCM adopted a system with TR = 31.67-36.32\;ms and TE = 1.39-1.41\;ms with 25-phase cardiac cycles.

\noindent\textbf{Evaluation Metrics.}
The concordance index~\cite{Harrell2001RegressionMS} (C-index) was used in our experiments as a metric that accounts for both continuous and interval-based survival prediction models. It quantifies the prediction effect based on the number of correct pairs. We have
\begin{equation}
\text{C-index}=\frac{\sum\limits_{i,j}\mathbb{I}(t_i<t_j)\mathbb{I}(r_i>r_j)\delta_i}{\sum\limits_{i,j}\mathbb{I}(t_i<t_j)\delta_i},
\end{equation}
where $\delta_i$ indicates event occurrence, $r_i=\mathbf{\beta}^\mathrm{T}\mathrm{x}_{\mathrm{fused}}^{i}$ is the risk score.

\noindent\textbf{Implementation Details.}
The proposed model was developed utilizing the PyTorch framework and trained on a single NVIDIA-H100 GPU with CUDA 12.2. The optimization process utilized the Adam optimizer~\cite{Kingma2014AdamAM}, with a learning rate of $5\times 10^{-5}$ and weight decay set to 
$1\times 10^{-5}$. A batch size of 16 was employed, and training was conducted over 50 epochs with the StepLR scheduler. To prevent overfitting, a penalizer was applied with values of $10^{-4}, 10^{-2}$, and $10^{-2}$, with feature correlation thresholds of 0.7, 0.9, and 0.7 for the RJCCM, AZCCM, and TJCCM datasets, respectively. The 4D image inputs were resized to a resolution of \( 24 \times 24 \times 96 \times 96 \), where the dimensions correspond to depth, frame, height, and width.

\subsection{Experimental Results}

\begin{table}[t]
\centering
\caption{Performance comparison across three datasets (Metric: C-index$\uparrow$ (\textit{p}-value$\downarrow$)). The radiomics features extracted using PyRadiomics v3.1.0~\cite{vanGriethuysen2017ComputationalRS} serve as substitutes of images for non-imaging models like DeepSurv and DSM.}\label{tab1}

\begin{tabular}{ccccccc}
\hline
Model &  EHR & Img   & Radiomics &  RJCCM &  AZCCM & TJCCM\\
\hline
CoxPH~\cite{Cox2018CoxPH} & \checkmark  & -  & - & 0.638 (.259) & 0.745 (.002) & 0.562 (.212)\\

DeepSurv~\cite{Katzman2016DeepSurvPT} & \checkmark & - & \checkmark & 0.608 (.120)& 0.618 (.109)& 0.623 (.088) \\

DSM~\cite{Nagpal2021DSM} & \checkmark & - & \checkmark  & 0.690 (.010) & 0.632 (.191) & 0.746 (.201) \\

SurvRNC~\cite{Sae2024SurvRNC} & \checkmark & \checkmark  &  - & 0.731 (.014) & 0.739 (.099) & 0.648 (.064)\\

Sparse BagNet~\cite{Ger2024Interpretablebydesign}  &  -&  \checkmark & -& 0.568 (.109)  & 0.715 (.008) & 0.545 (.244)  \\

CTSL (Ours) & \checkmark  & \checkmark & -  &  \textbf{0.788} (.074) & \textbf{0.826} (.036) & \textbf{0.863} (.029)\\
\hline
\end{tabular}
\label{tab:t1}
\end{table}

Table~\ref{tab:t1} presents comparative results between classical and SOTA models. Our proposed CTSL demonstrates robust risk prediction capabilities across three cohorts, with C-index values of 0.788, 0.826, and 0.863, respectively, outperforming both the clinical-dependent model cluster, including CoxPH, DeepSurv, DSM, as well as the SOTA models SurvRNC and Sparse BagNet.

\begin{figure}[t]
    \centering
    \includegraphics[width=0.85\linewidth]{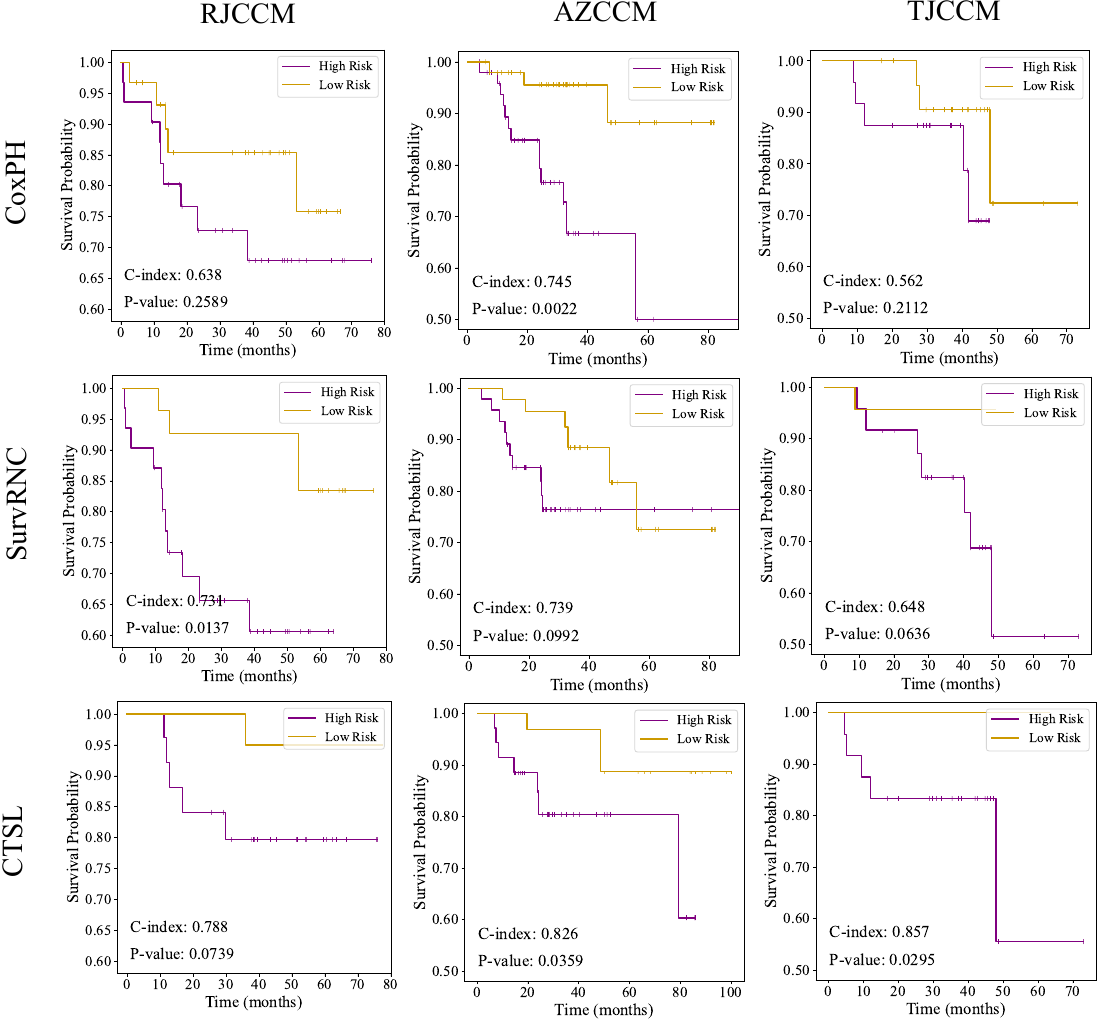}
    \caption{Kaplan-Meier~\cite{Kaplan1958NonparametricEF} analysis comparing risk stratification performance. Curves contrast our model against the clinical gold-standard CoxPH and SOTA SurvRNC baseline. Patients were stratified into high/low-risk groups by median predicted risk scores.}
    \label{fig:KM}
\end{figure}
Detailed Kaplan-Meier survival analysis is provided in Fig.~\ref{fig:KM}, with \textit{p}-values incorporated from the log-rank test to highlight statistical significance. Comparisons include the clinical gold-standard CoxPH and the multimodal SurvRNC (top-performing baseline). CTSL achieves the lowest \textit{p}-values on average, with complete separation between high- and low-risk groups while no intersection of the curve is observed. Besides, increasingly pronounced prognostic differentiation is detected over time.

\begin{figure}[htbp]
    \centering
    \includegraphics[width=0.98\linewidth]{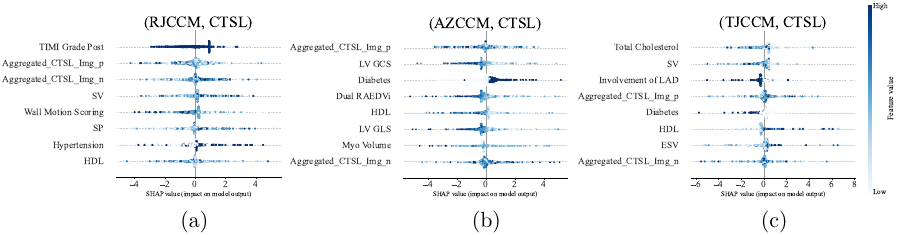}
    \caption{SHAP-based interpretability analysis (a: RJCCM, b: AZCCM, c: TJCCM). Top-8 features are displayed for each dataset. The top-5 most prognostically influential imaging biomarkers with positive (Aggregated\_CTSL\_Img\_p) and negative (Aggregated\_CTSL\_Img\_n) contributions are aggregated, respectively.}
    \label{fig:shap}
\end{figure}

\noindent\textbf{Interpretable Comparison.}
Clinically, high-density lipoprotein (HDL) levels, diabetes status, and stroke volume (SV)  emerged as key clinical determinants of MACEs, exhibiting cross-center stability in  contribution magnitude as Fig.~\ref{fig:shap} shows. The CTSL-derived imaging biomarkers revealed myocardial motion signatures with superior predictive value. Notably, cine motion-driven features, including wall motion scoring, end-systolic volume (ESV), and dual-chamber right atrial end-diastolic volume index (Dual RAEDVi),  in synergy with imaging data, collectively demonstrated significant risk stratification power.

\noindent\textbf{Ablation Study.}
To evaluate the robustness of our framework, we design ablation experiments at three levels: (1) Model CTSL, which is obtained through the complete workflow; (2) Model Uniformer(Distilled), whose representation aggregated from motion queries directly without employing the discrete spatiotemporal codebook for refinement; (3) Model Uniformer(ImageNet), which does not undergo distillation or codebook discretization, and instead relies solely on pre-trained ImageNet~\cite{Deng2009ImageNet} weights as a feature extractor.

\begin{table}[htbp]
\centering
\caption{Results of ablation studies (Metric: C-index).}\label{tab2}
\begin{tabular}{cccccc}
\hline
Model &  Distillation & Quantization & RJCCM &  AZCCM & TJCCM\\
\hline
Uniformer (ImageNet) & -  & -  & 0.608 & 0.661 & 0.621\\

Uniformer (Distilled) & \checkmark & - & \textbf{0.842} & 0.754 & 0.648\\

CTSL & \checkmark & \checkmark & 0.788  & \textbf{0.826} & \textbf{0.863} \\

\hline
\end{tabular}
\label{tab:t2}
\end{table}
As evidenced in Table~\ref{tab:t2}, our Stage I distillation framework demonstrates superior performance over natural image pre-trained counterparts through the synergistic integration of multi-view cardiac dynamics (mean $\Delta $C-index: +0.118 vs. baselines). While achieving marginally lower performance in cohort RJCCM, Stage II's disentangled spatiotemporal representations achieve statistically an overall performance improvement (mean $\Delta $C-index: +0.078 vs. Stage I). This cross-cohort consistency quantitatively validates the robustness of our latent space learning paradigm against anatomical variability.

\section{Conclusion}
This study introduces a self-supervised framework for non-contrast cardiac risk prediction, integrating motion-aware model distillation with codebook-based spatiotemporal disentanglement. By eliminating manual annotations, our approach effectively captures intrinsic myocardial dynamics. Experimental results demonstrate that CTSL not only enhances prognostic accuracy but also improves model interpretability by transforming raw 4D Cine sequences into relevant image biomarkers. These findings highlight the potential of routine imaging for risk stratification, laying the groundwork for future advancements in personalized therapeutic planning through dynamic motion trajectory modeling.

\begin{credits}
\subsubsection{\ackname} 
This work was supported by the Shanghai Artificial Intelligence Laboratory, the National Natural Science Foundation of China (Grant Nos. 82171884 and 82471931), and the Shanghai Municipal Commission of Science and Technology Medical Innovation Research Special Project (Grant No. 23Y11906900).

\subsubsection{\discintname}
The authors have no competing interests to declare
 that are relevant to the content of this article.
\end{credits}

\bibliographystyle{splncs04}
\bibliography{references}

\end{document}